\documentclass[a4paper, 10pt, conference]{ieeeconf}      
\usepackage{FG2017}

\FGfinalcopy 

\IEEEoverridecommandlockouts                              
\usepackage{graphics} 
\usepackage{graphicx}
\graphicspath{ {figures/}}

\title{\LARGE \bf{Deep Convolutional Neural Network Features and the Original Image }}

\author{\parbox{16cm}{\centering
    {\large Connor J. Parde$^1$ and Carlos Castillo$^2$ and Matthew Q. Hill$^1$ and Y. Ivette Colon$^1$ and Swami Sankaranarayanan$^2$ and Jun-Cheng Chen$^2$ and Alice J. O'Toole$^1$  }\\
    {\normalsize
    $^1$ School of Behavioral and Brain Sciences, The University of Texas at Dallas, USA\\
    $^2$ Department of Electrical Engineering, University of Maryland, College Park, USA}}
}

\begin{document}

\maketitle

\begin{abstract}
Face recognition algorithms based on deep convolutional neural networks (DCNNs) have made progress on the task of 
recognizing faces in unconstrained viewing conditions. These networks operate with compact 
feature-based face representations derived from learning a very large number of face images.
While the learned feature sets produced by DCNNs can be highly robust to changes in viewpoint, illumination, 
and appearance, little is known about the nature of the face code that emerges at the top level of such networks.
We analyzed the DCNN features produced by two recent face recognition algorithms.
In the first set of experiments,  we used the
top-level features from the DCNNs as input into linear classifiers 
aimed at predicting metadata about the images.
The results 
showed that the DCNN features contained surprisingly accurate information about the
yaw and pitch of a face, and about whether the input face came from a still image or a video frame.
In the second set of experiments, we measured the extent to which individual DCNN features 
operated in a view-dependent or view-invariant manner for different identities. We found that view-dependent coding
was a characteristic of the \textit{identities} rather than the DCNN features--with some identities coded consistently in a view-dependent
way and others in a view-independent way. In our third analysis, we visualized the DCNN feature space for 24,000+ images
of 500 identities. Images in the center of the space were uniformly of low quality (e.g., extreme views, face occlusion, 
poor contrast, low resolution). Image quality increased monotonically as a function of distance from the origin. This result suggests that image quality information is available in the DCNN features, such that consistently average feature values reflect coding failures that reliably indicate poor or unusable images.
Combined, the results offer insight into the coding mechanisms that support robust representation of faces in DCNNs.  
\end{abstract}

\section{INTRODUCTION}

Face recognition algorithms based on convolutional neural networks and deep learning show considerable robustness to changes in imaging parameters (e.g., pose, illumination, and resolution) and facial appearance (e.g., expression, eyewear). This robustness accounts for the impressive gains made by CNNs on the problem of unconstrained face recogniton \cite{taigman2014deepface,parkhi2015deep,simonyan2013fisher,schroff2015facenet,DBLP:journals/corr/HuYYKCLH15,huang2012learning}.   
Performance on datasets such as LFW \cite{LFWTech,Kumar:2009cs}, IJB-A \cite{JC15,SwamiBTAS}, and Mega-Face \cite{miller2015megaface} offer evidence that  face recognition by machines can, in some cases, approach human performance \cite{taigman2014deepface}. Indeed, human recognition of familiar faces (e.g., friends, family) operates in highly unconstrained environments and over changes in appearance and age that can span decades. This kind of performance remains a goal of automated face recognition systems.

Although humans remain a proof-of-principle that highly invariant face recognition is possible, the underlying nature of the face representation that supports invariance in humans is poorly understood. The nature of the representation captured in DCNN features is similarly elusive.
The goal of this paper is to characterize the features that emerge in a DCNN trained for face recognition so as to better understand why they are robust to yaw, pitch, and media type (still image or video frame). The approach we take is to first examine the extent to which the ``robust" feature sets that emerge in a CNN {\it retain} information about the original images.  
As we will see, DCNNs that show considerable robustness to pose and media type retain
detailed information about the images they encode at the deepest and most compact level of the network. Second, we explore the view-dependency and media-dependency characteristics of DCNN features. Third, we examine cues pertaining to image quality 
within the structure of the feature space. 

\section{Background and Problem}

The problem of image-invariant face perception has been studied for decades in both computer vision \cite{SungPoggio98} and psychology. 
Traditionally, two classes of models have been considered: a.) representations that capture 3D facial structure and b.) representations based on collections of 2D, image-based views of faces. The former can enable specification of appearance across arbitrary affine and non-affine transformations.
The latter can show invariance in any given instance via interpolation to image representations taken in conditions similar to that of the probe image. Notably, this requires ``experience" with enough diverse views to be successful across a range of possible probes. Active appearance models \cite{BlanzVetter99} comprise an intermediary class, which relies on class-based knowledge of faces, including 3D structure and reflectance-map information for many examples. Although these models can achieve impressive performance in computer graphics representations made from single images, they are not practical for face recognition as they are computationally intense and require high quality, 3D data on diverse classes of faces. 

The recent gains made in face recognition can be tied both to the computational power of DCNNs and to the quality and quantity of the training data now available from web-scraping. In theory, the goal of a DCNN is to develop an invariant representation of an individual's face through exposure to a wide variety of images showing that person in different settings, with different poses, and in images that vary in quality. Given enough data, it is expected that the network will be able to learn a representation of an individual that does not rely on these non-static, image-level attributes. Instead, the intent is that the learned features represent the invariant information in a face that makes the face unique.

The fact that DCNNs support robust recognition across image transformation does not preclude the possibility that the features used to code faces in these networks also retain information about the image properties. Rather, DCNNs may succeed across appearance-related and image-related variation
by incorporating both identity and image parameters into the face codes. This  code may support the separation of image and identity
for identity verification. This separation may ultimately be achieved at a post-DCNN stage via another type of classifier that operates on image or person representations extracted from the deepest, most compact layer of the DCNN.

\begin{figure}[t]
\includegraphics[width=\columnwidth,height=6.5cm]{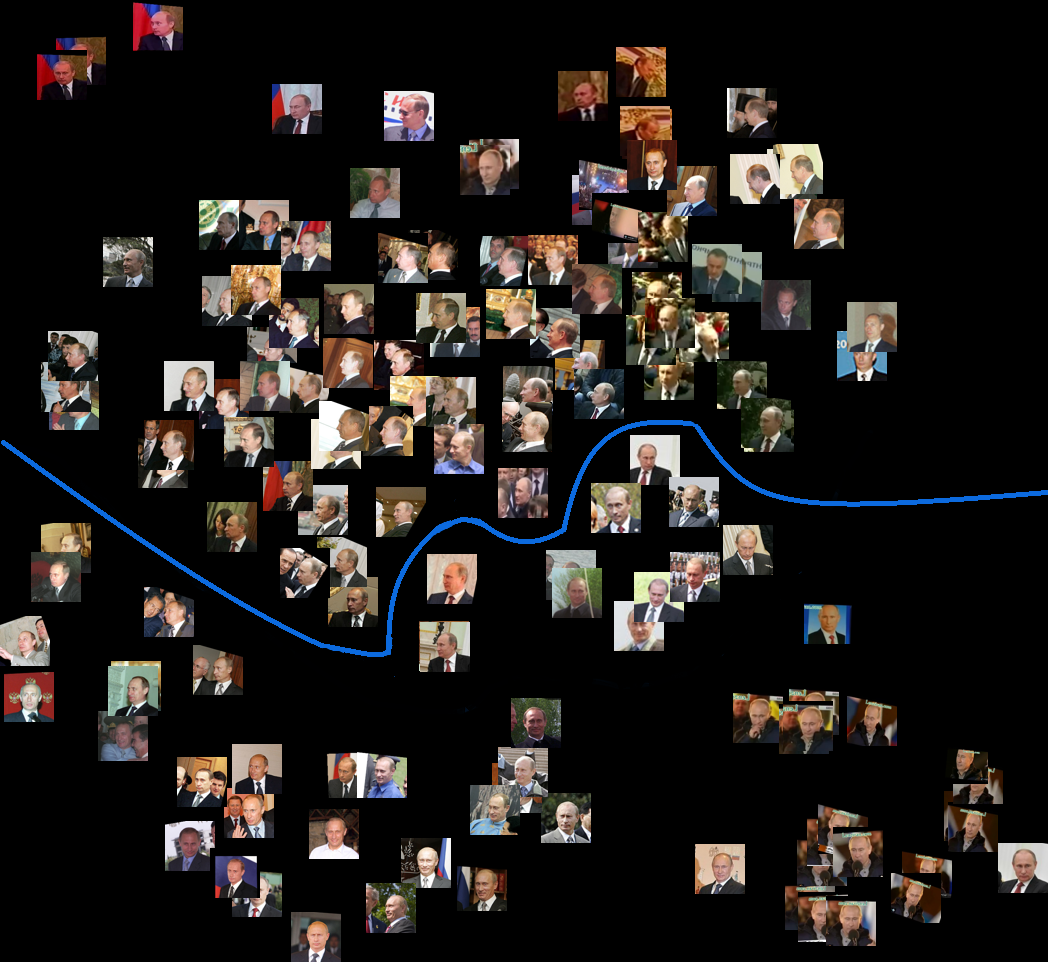}
\vskip .05cm
\includegraphics[width=\columnwidth,height=6.5cm]{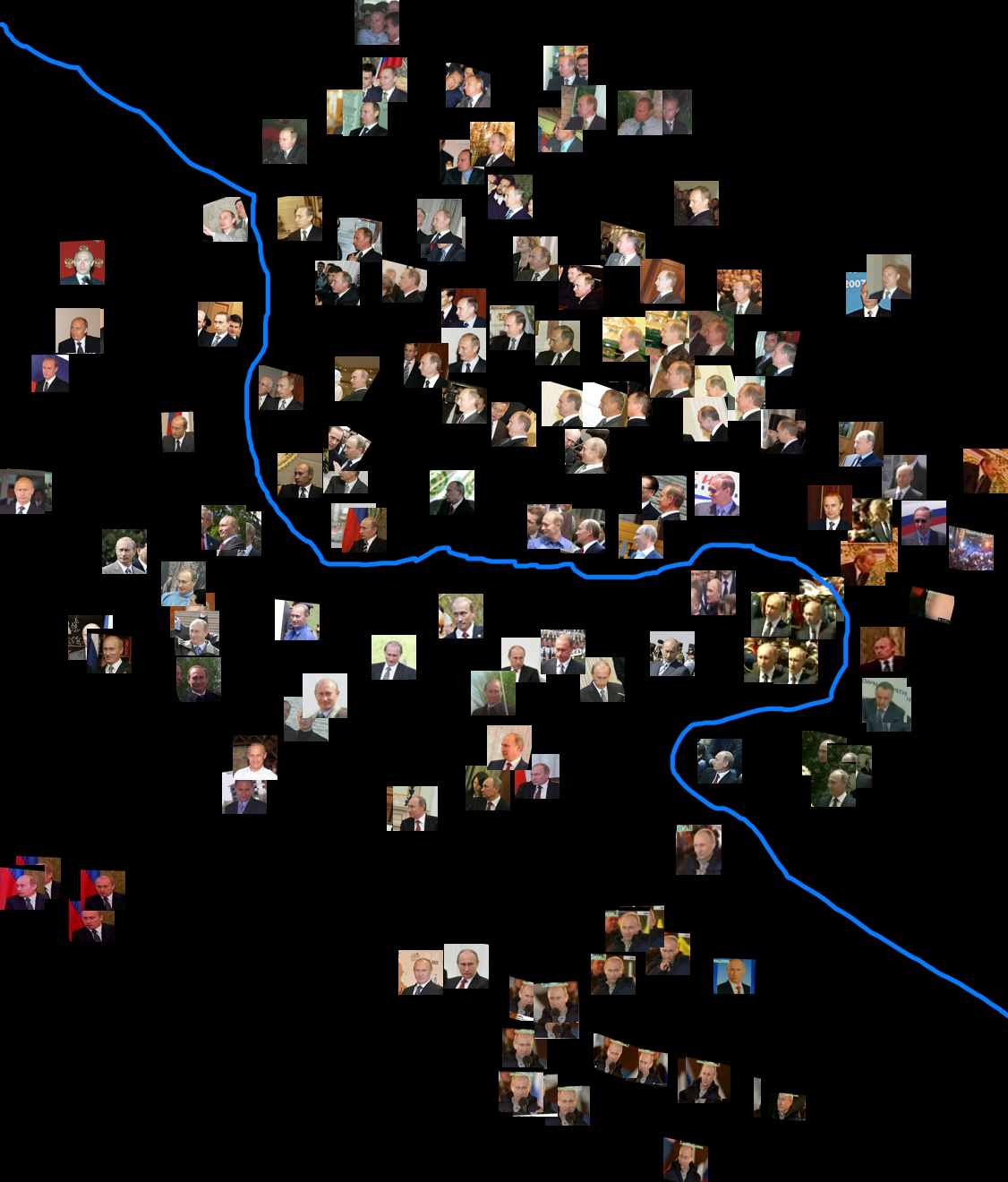}
\caption{These figures show the t-SNE visualization of the top level DCNN features for 140 images of Vladimir Putin from the Janus CS2 dataset.
The visualizations are based on the 320 top-level DCNN features from Network A \cite{JC15} (top) and the 512 top-level DCNN features from Network B \cite{SwamiBTAS}.}
\label{fig:Vladimir}
\end{figure}

The motivation for the present work came from visualizing the way {\it single identities} cluster in a low-dimensional space derived from the top-level features produced by two recent DCNNs \cite{JC15,SwamiBTAS}. These DCNNs were developed to work on the Janus CS2 dataset, an expanded version of the IJB-A dataset \cite{IIJBA_dataset_2016}. We describe the architecture of the two DCNNs in detail in the methods section. For present purposes, this visualization was done by applying t-Distributed Stochastic Neighbor Embedding (t-SNE) \cite{tSNE} to the top level features of each network. t-SNE is a dimensionality reduction technique that uses stochastic probability methods to preserve the high-dimensional Euclidean distances between data points while embedding them in a low-dimensional space. We visualized single identities that had large numbers of images available in the Janus CS2 dataset. Figure \ref{fig:Vladimir} shows the t-SNE space for the top level features of 140 CS2 images of Vladimir Putin, extracted from the two DCNNs. Both plots exhibit roughly separable clusters of profile and frontal images of the subject. The blue curves were hand-drawn onto the visualizations to indicate the position of an approximate border. 

The clustering suggests that the top-level features produced by both of these DCNN networks preserve salient, view-related information captured in the original image, while still clustering by identity. More generally, this suggests that DCNNs contain a deeper-than-expected representation of the original image in their top-level features. Notably, the clustered images of Putin still varied substantially 
in other appearance- and image-based attributes (e.g., age, illumination). 

In what follows, we quantify the clustering behavior of image-based attributes in these two DCNNs. 
This paper is organized as follows. In Section III, we present the networks and the datasets analyzed.
In Section IV we use the top-level features of the DCNNs as input into linear classifiers aimed at predicting metadata about the images including yaw, pitch, and media type (still image or video). In Section V, we analyzed the extent to which top-level features operate invariantly across viewpoint and media type. In Section VI, we examine the top-level feature space of image representations in the context of image quality.

\section{General Methods}

\subsection{Description of DCNN's}

We analyzed the feature-space produced by two DCNNs (Network A, \cite{JC15}; Network B, \cite{SwamiBTAS}) using the Janus CS2 dataset. Both networks approach the problem by constructing a feature-based representation of all input images using a DCNN. Full details on these networks are available elsewhere. For present purposes, we discuss only the training of these two networks since our analysis focuses only on the top-level features produced by either network.

The base architectures of the DCNNs appear in Tables \ref{table_JC} and \ref{table_Swami}. In both networks, parametric ReLU (PReLU) were used as the activation function. In Network A, a learned feature space was developed from scratch and produced a 320-dimensional feature vector for each input image. The second network (Network B) builds upon the AlexNet model \cite{krizhevsky2012imagenet} and assigns each input image a 512-dimensional feature vector. At its lower levels, Network B initially assigns weights based on the values generated by AlexNet and then trains its higher layers using the CASIA-Webface database. Network A also uses CASIA-Webface for training and does so for all layers in the network.

\begin{table}
\caption{Network A}
\label{table_JC}
\begin{center}
\begin{tabular}{|c||c||c||c||c||c|}
\hline
Name  & Filter Size/Stride & Output  & Parameters \\
\hline
conv11 & 3x3x1/1 & 100x100x32 & .28K    \\
conv12 & 3x3x32/1 & 100x100x64 & 18K    \\
pool1 & 2x2/2 & 50x50x64 &     \\
conv21 & 3x3x64/1 & 50x50x64 & 36K    \\
conv22 & 3x3x64/1 & 50x50x128 & 72K    \\
pool2 & 2x2/2 & 25x25x128 &     \\
conv31 & 3x3x128/1 & 25x25x96 & 108K    \\
conv32 & 3x3x96/1 & 25x25x192 & 162K    \\
pool3 & 2x2/2 & 13x13x192 &     \\
conv41 & 3x3x192/1 & 13x13x128 & 216K    \\
conv42 & 3x3x128/1 & 13x13x256 & 288K    \\
pool4 & 2x2/2 & 7x7x256 &     \\
conv51 & 3x3x256/1 & 7x7x160 & 360K    \\
conv52 & 3x3x160/1 & 7x7x320 & 450K    \\
pool5 & 7x7/1 & 1x1x320 &     \\
dropout (40\%) & & 1x1x320 &     \\
fc6 & & 10548 & 3296K   \\
softmax cost & & 10548 &     \\
total & & & 5006K    \\
\hline
\end{tabular}
\end{center}
\end{table}

\begin{table}
\caption{Network B}
\label{table_Swami}
\begin{center}
\begin{tabular}{|c||c||c|}
\hline
Layer & Kernel Size/Stride & Parameters \\
\hline
conv1 & 11 x 11/4 & 35K \\
pool1 & 3 x  3/2 &  \\
conv2 & 5 x 5/2 & 614K \\
pool1 & 3 x  3/2 &  \\
conv3 & 3 x 3/2 & 885K \\
conv4 & 3 x 3/2 & 1.3M \\
conv5 & 3 x 3/1 & 2.3M \\
conv6 & 3 x 3/1 & 2.3M \\
conv7 & 3 x 3/1 & 2.3M \\
pool7 & 6 x 6/2 &  \\
fc6 & 1024 & 18.8M \\
fc7 & 512 & 524K \\
fc8 & 10548 & 10.8M \\
Softmax Loss &  & Total 39.8M \\
\hline
\end{tabular}
\end{center}
\end{table}

\subsection{CS2 Dataset}
The images were sourced from the CS2 dataset. The dataset includes 25,800 images of 500 subjects.
CS2 is an expanded version of the IARPA Janus Benchmark A (IJB-A) \cite{IIJBA_dataset_2016}, a publicly available ``media in the wild" dataset. Some key features of the IJB-A dataset include: full pose variation, a mix of images and videos, and a wider demographic variation of subjects than is available in the LFW dataset. The dataset was developed using 1,501,267 crowd sourced annotations. Baseline accuracies for both face detection and face recognition from commercial and open source algorithms are available in \cite{IIJBA_dataset_2016}.

The original IJB-A included metadata from crowd-sourcing. Here we used  
metadata provided by the Hyperface system described in \cite{ranjan2016hyperface}. The Hyperface system provides key-point locations to aid in face detection, as well as estimated measurements of face pose (yaw, pitch, and roll).

Of the 25,800 items in the CS2 dataset, we omitted 1,298 items from our analysis. This was due to either Network A's or Network B's inability to compute features for one of these images, or Hyperface's inability to compute the pose of the subject within an image. This left us with 24,502 items that could be considered when training classifiers to predict each metadata attribute of interest.

\section{Predicting Image-Related Metadata from the DCNN Features}
Each experiment described in this section consisted of a bootstrap test of metadata prediction based on the top-level feature encodings from Network A and B. Predictions were computed using a linear discriminant analysis (LDA) classifier, with 20 iterations of the bootstrap test for each metadata attribute. For each iteration, we randomly selected 18,000 items to use as training data. We tested the classifier on the remaining 6,502 items.  The results reported on prediction
accuracy are averaged across the 20 bootstrapped iterations. 

\subsection{Predicting Yaw}

The yaw values provided by the Hyperface system for the CS2 dataset describe the yaw angle of the face in an image, measured in degrees, and varying from -90 (left profile) to +90 (right profile), with 0 indicating a frontal pose. For both networks, pre-processing steps were performed to mirror all left-facing images, thereby limiting the yaw range to only positive values. Therefore, we used the absolute value of the yaw scores provided by Hyperface as output for the classifier.
In each bootstrap iteration, a classifier was trained to predict the Hyperface
yaw values from the DCNN features. Prediction accuracies for both Networks A and B 
appear in Table \ref{yaw_pitch_table} and are surprisingly high, to within less than 9 degrees, and are  consistent across bootstrap iterations.

\begin{table}
\caption{Yaw and Pitch Predication Accuracy}
\label{yaw_pitch_table}
\begin{center}
\begin{tabular}{|c||c||c|}
\hline
 Network & Yaw & Pitch  \\
\hline
A & +/-8.06 degs. ({\it sd.} 0.078) & 77.0\% correct  \\
\hline
B & +/-8.59 degs. ({\it sd.}  0.071) & 71.5\% correct  \\
\hline
\end{tabular}
\end{center}
\end{table}

\subsection{Predicting Pitch}

Pitch estimates for the dataset were provided by Hyperface and measured in degrees. A positive pitch score indicates an upward looking face, a negative pitch indicates a downward looking face, and a score of 0 indicates a face looking directly at the camera. Given that the majority of images in the CS2 dataset showed faces with a relatively centered pitch, pitch was coded categorically for this experiment as {\it centered} and {\it deviating} pitch. Centered pitch was defined as all values between -8 and +8 degrees. Deviating pitch was defined as all values outside of the centered range. 

Using the DCNN features as input, we predicted whether each image in the CS2 data set showed a face with centered  pitch or deviating pitch. Predictions on the test data were continuous values from 0 (centered) to 1 (deviating). These were rounded to the nearest integer (0 or 1) to obtain the prediction values. The results appear in Table III, reported as percent correct. As with yaw, pitch category prediction from the DCNN features was unexpectedly accurate (77.0\%  and 71.5\% correct) for Networks A and B, respectively.

\subsection{Predicting Media Type}

The media type of each image was provided in the CS2 dataset. Each image 
originated as either a still photograph or a video frame. 
An image's media type might be considered a proxy-measure for some aspects 
of image quality. In the CS2 dataset, the images that originated as still photographs were typically better illuminated and had higher resolution. The images that originated as video frames tended to come from lower-quality data sources such as CCTV footage.

We assigned a score of 1 to all images in the CS2 dataset that originated as still photographs, and a score of 0 to all images that originated as video frames. We then applied the bootstrapped classification method to predict media type from the CNN features. The predictions for our test data were continuous values from 0 to 1. These were rounded to the nearest integer (0 or 1) to obtain the prediction values. The results appear in Table \ref{media_type_table}, reported as percent correct. Predictions using the DCNN features were highly accurate and consistent for both networks.

\begin{table}
\caption{Media Type}
\label{media_type_table}
\begin{center}
\begin{tabular}{|c||c||c|}
\hline
Network & Media Type\\
\hline
A & 87.1\% (sd. 0.004)\\
\hline
B &  93.3 \% (sd. 0.002)\\
\hline
\end{tabular}
\end{center}
\end{table}

\begin{figure}[h]
\includegraphics[width=\columnwidth]{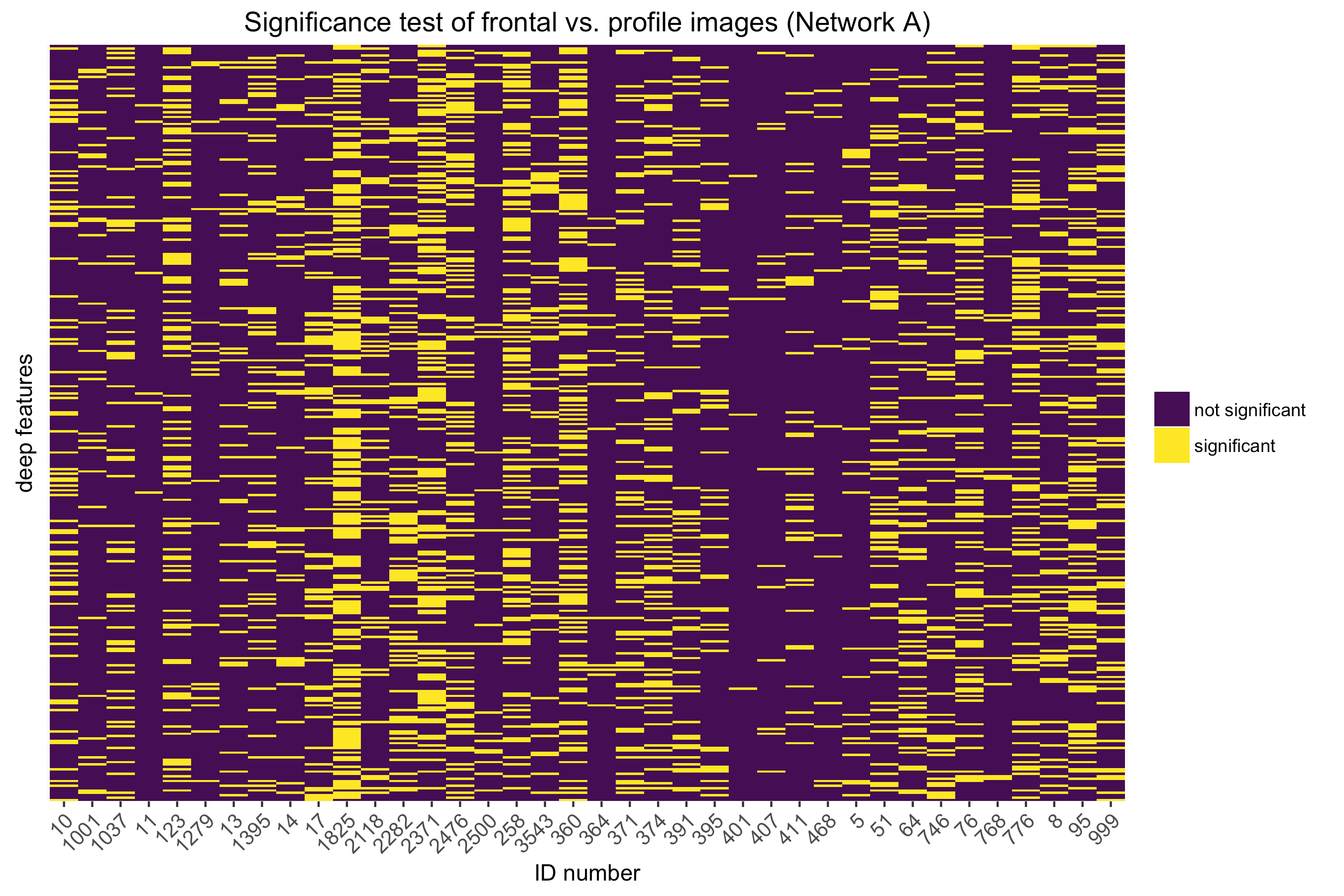}
\includegraphics[width=\columnwidth]{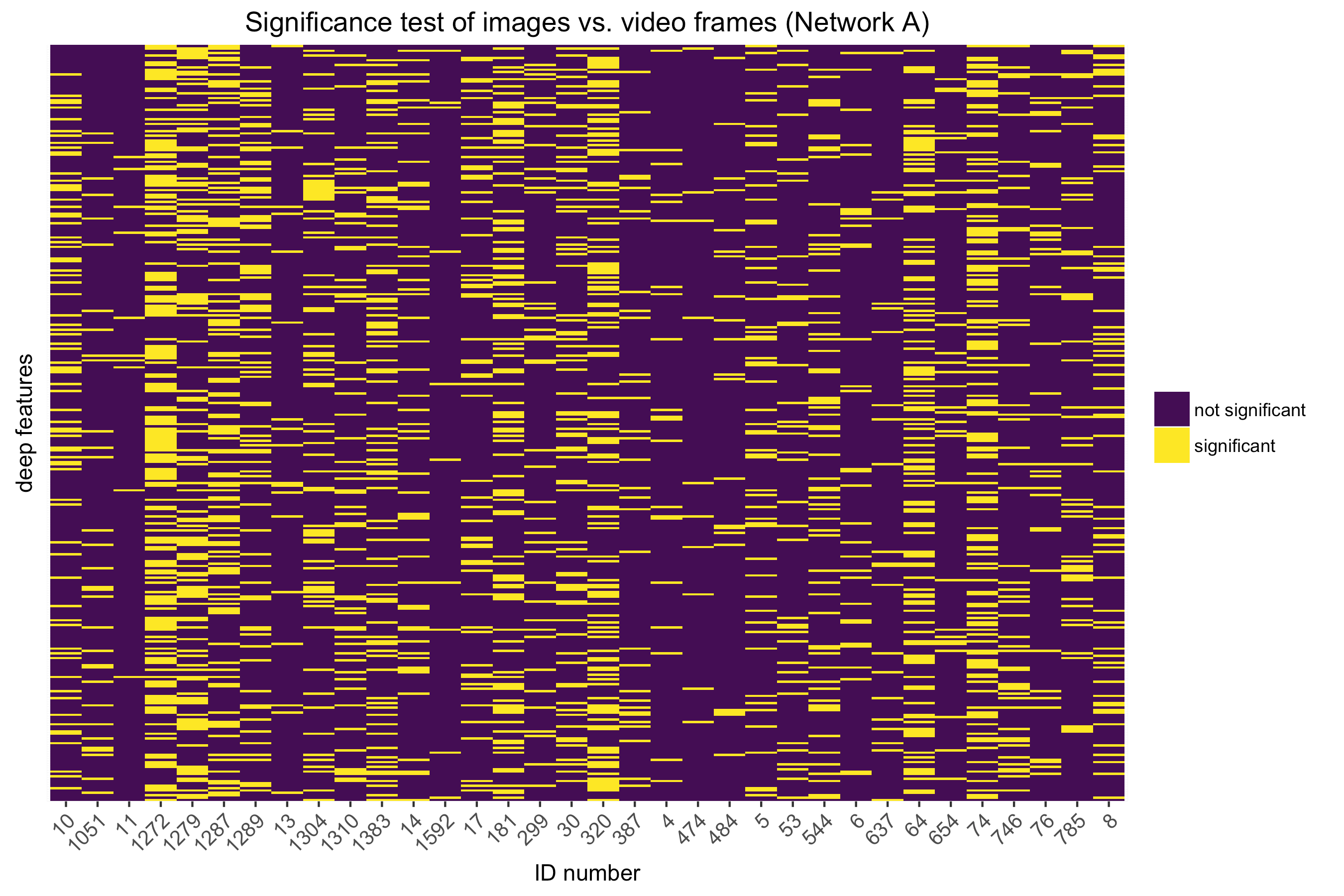}
\caption{Heat map illustration of view-dependent DCNN features for Network A displayed for each identity in the database with at least 20 frontal and 20 profile images (top). Heat map illustration of the quality-dependent top-level DCNN features for each identity in the database with at least 20 still images and 20 video frames (bottom).}
\label{fig:heatmaps}
\end{figure}

\subsection{Interim summary}
The classification experiments showed that metadata from individual images, including yaw, pitch, and media type, was available in the top level DCNN features of both networks. 

In the next section the goal was to analyze the extent to which individual features operate invariantly, or at least robustly, across 
pose and media type. 

\begin{figure}[h]
\center{
\includegraphics[height=3.5cm]{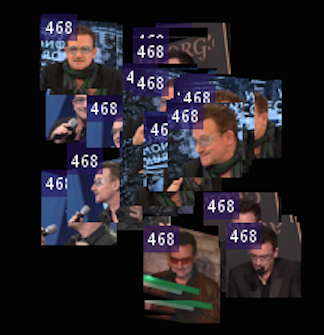}
\includegraphics[height=3.5cm]{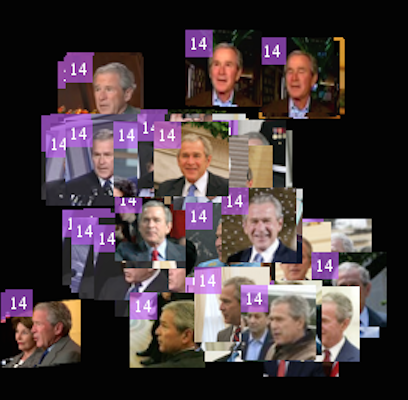}}

\caption{Image clusters of two individuals (Bono and G. Bush) who were both coded with a majority of view-independent features (312 and 289 of 320 respectively). These clusters show mixed viewpoints aligned closely, which may correspond to distinctive features (e.g. Bono's sunglasses) that are easy to detect across variable views.  }
\label{fig:bono_bush}
\end{figure}

\begin{figure}[h]
\center{
\includegraphics[width=7cm]{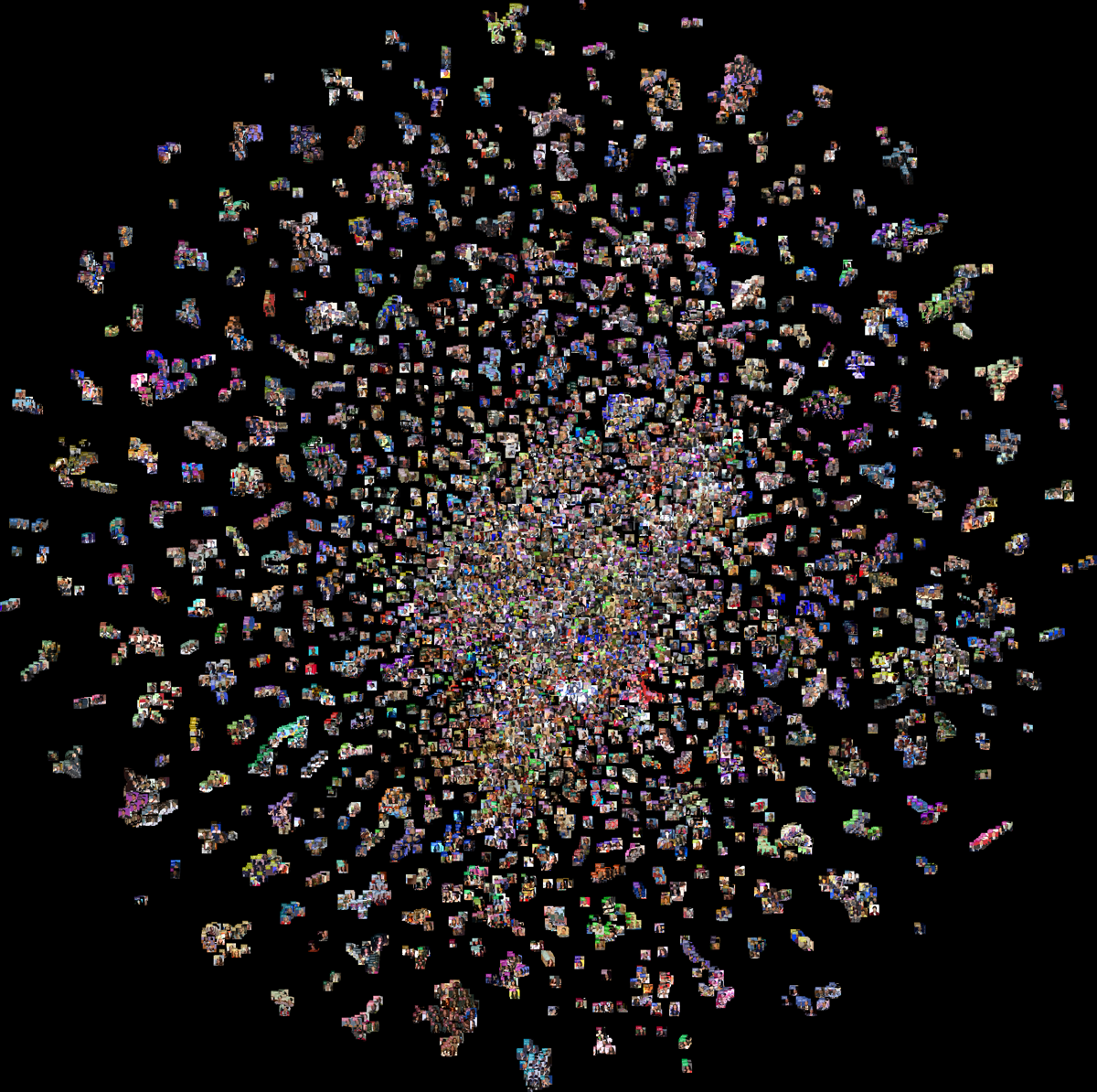}\\
\vskip .05cm
\includegraphics[width=7cm]{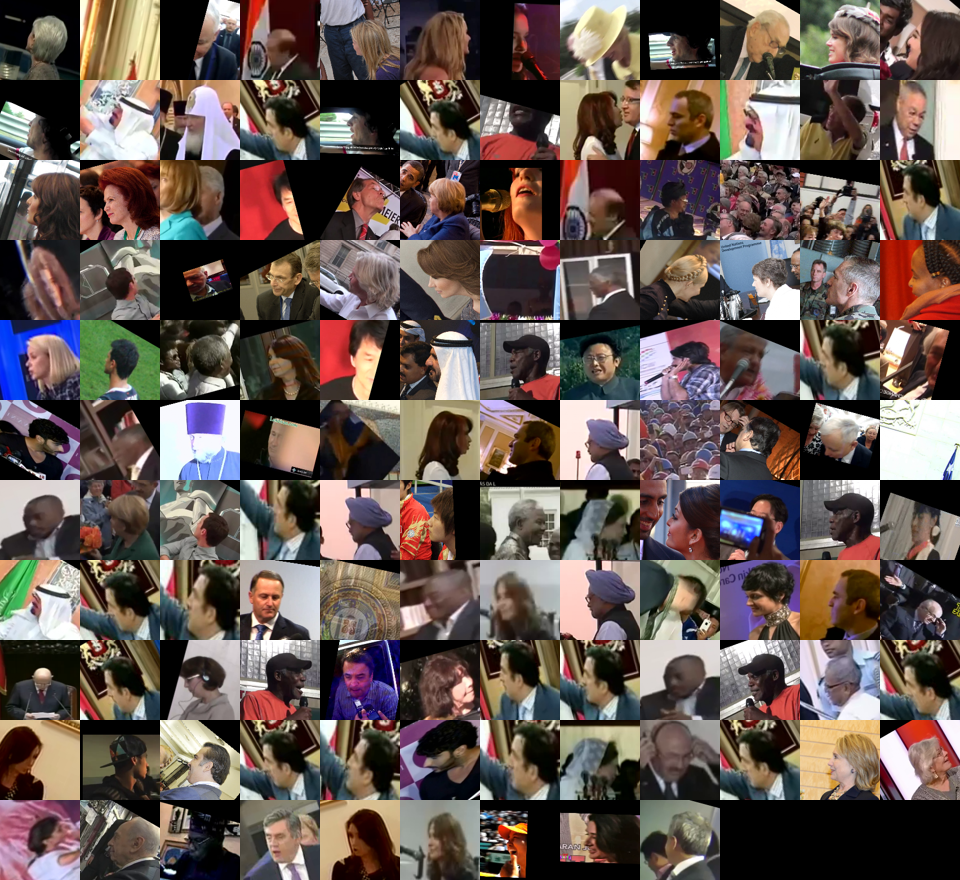}\\
}
\caption{Results of t-SNE applied to the DCNN top level features of Network A for all 24,502 images (top). An array of the 129 images closest to the center of the space (0.05\%) in Network A. The upper-left image is the image closest to the center, and each image's distance from the center grows as you progress across the rows (bottom).}
\label{fig:whole_space}
\end{figure}

\section{CNN Features and Invariance: Are features invariant or are people invariant?}
\subsection{View (In)variance Coding}
Beginning with view, we developed an index of feature robustness across frontal and profile poses. We approached the problem as follows. First, we sub-selected identities in the database ({\it n} = 38) for which there were at least 20 frontal images and 20 profile images. Second, within each of these identities, for each of the 320 DCNN features in Network A, we computed a {\it t}-test to determine
whether the feature's values from frontal images of that individual differed significantly from the feature's values from profile images.  We set the alpha level
for statistical significance at 0.000156\footnote{This is a two-tailed alpha level of 0.05, Bonferroni corrected for 320 multiple comparisons.}. The resultant {\it p}-values act as an of index of feature invariance for an individual. The results of this analysis are displayed in the top panel heat map in Figure \ref{fig:heatmaps} and are surprising. In the figure, individual identities are displayed across columns and individual features are displayed across rows. We anticipated that individual features would consistently code identities in either a view-dependent or view-invariant way. This would have produced horizontal bands in the heat map, suggesting the consistency of a feature across identities. Instead we found the inverse, with individual identities being coded in either a view-dependent or view-invariant way across features. This is indicated by the vertical lines of significant features in the heat map. More formally, the percentage of features that differentiated faces by viewpoint for an individual was as high as 55.31\%. Individual features did not consistently code in a view-dependent or view-independent manner.

To interpret these heat maps, we visualized the most- and least-differentiated identities by selecting the most strongly banded columns from the heat map. Two examples of non-differentiated identities appear in Figure \ref{fig:bono_bush} and show Bono and Pres. George W. Bush. For Bono, 90.31\% of the 320 features were undifferentiated as a function of view; for Bush, 97.5 \% were undifferentiated. These clusters show mixed viewpoints aligned closely--possibly reflecting the presence of distinctive identity features that are easy to detect in any view (e.g. Bono's oddly tinted
sunglasses). In visualizing identities with the most differentiated features, however, many subjects show strongly separated clusters, each of which shows a small range of similar views. This latter pattern resembles what we saw in Figure \ref{fig:Vladimir} for Vladimir Putin. The main point, though, is that it is the {\it identity} that determines whether the features will operate in a view-dependent or view-invariant manner. Some identities are marked most strongly by characteristics which are static across shifts in pose, while others are marked by the way certain traits appear when seen from different viewing angles.

\begin{figure}[t]
\includegraphics[width=\columnwidth]{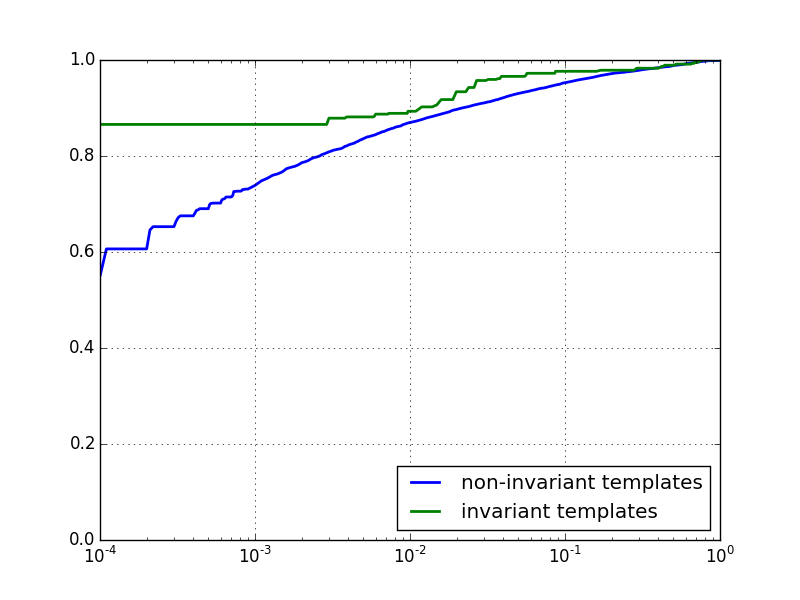}
\caption{Identity verification performance of Network A for template pairs where both identities are coded view-invariantly versus  for
all other template pairs. View-invariance of an identity  is characterized by feature values across its images 
that do not dissociate for frontal and profile views.}
\label{fig:invgps}
\end{figure}

To determine the extent to which the nature of an identity code (view-invariant or non-invariant) affects performance in a face recognition
algorithm, we conducted the following experiment. We selected the 7  identities coded most invariantly over view-change. Next we compared
the performance of Network A on template comparisons comprised of pairs of these 7 identities against templates comprised of all other 
identity pairs. Note that a template is defined as a variably sized set of images and video frames of an individual identity. 
The contents of the templates were specified by the Janus protocol.  The results appear in Figure \ref{fig:invgps} and show a strong advantage
for recognizing identities that can be coded invariantly, over those in which feature values dissociate for  frontal and profile images.

\subsection{Media Type (In)variance Coding}
We repeated the same approach from the previous section to examine the way media type is coded across features and individuals, developing an index of feature robustness across still images and video frames. First, we sub-selected identities in the database ({\it n} = 34) for which there were at least 20 still images and 20 video frames. Second, within each of these 34 identities, for each of the 320 top-level DCNN features produced by Network A, we computed a {\it t}-test to determine whether a feature's value for still images of that individual differed significantly from that feature's value for video frames. We again set the alpha level for statistical significance at 0.000156. In this case, the {\it p}-values act as an index of the feature's invariance for coding an individual in a still photograph versus in a video frame. The results of this analysis are displayed in the heat map in Figure \ref{fig:heatmaps} (bottom panel) and echo what is seen in the heat map distinguishing frontal and profile views. Individual identities tend to be coded in either a media-dependent or media-independent manner.

\section{When DCNN Features Fail They Leave a Trail}

We returned to the use of t-SNE to visualize the feature spaces of our two recognition networks. This time, rather than analyzing the feature space for a single individual, we applied t-SNE to the DCNN top level features for all 24,502 images (see Figure 4, top). This was used as an exploratory analysis to help us visualize the DCNN feature space in more detail. The primary insight gained from this visualization is that the images located near the center appear to be of extremely poor ``quality'', where quality refers to a wide range of issues that would make the person in the image difficult to detect or identify. We therefore examined the images in order of closeness to the center of the raw
feature space, using the origin of the feature space as the center.
Figure \ref{fig:whole_space} (bottom) shows an array of the 129 images 
closest to the center of the space (0.05\%) in Network A, arranged across the rows and starting from the image closest to the center. As seen in the array, the images closest to the center of the feature space are affected by a range of problems including extreme views, strong occlusion, blurring, distortion, and lack of an identifiable face.

\begin{figure}[h!]
\centering
\includegraphics[width=6.0cm]{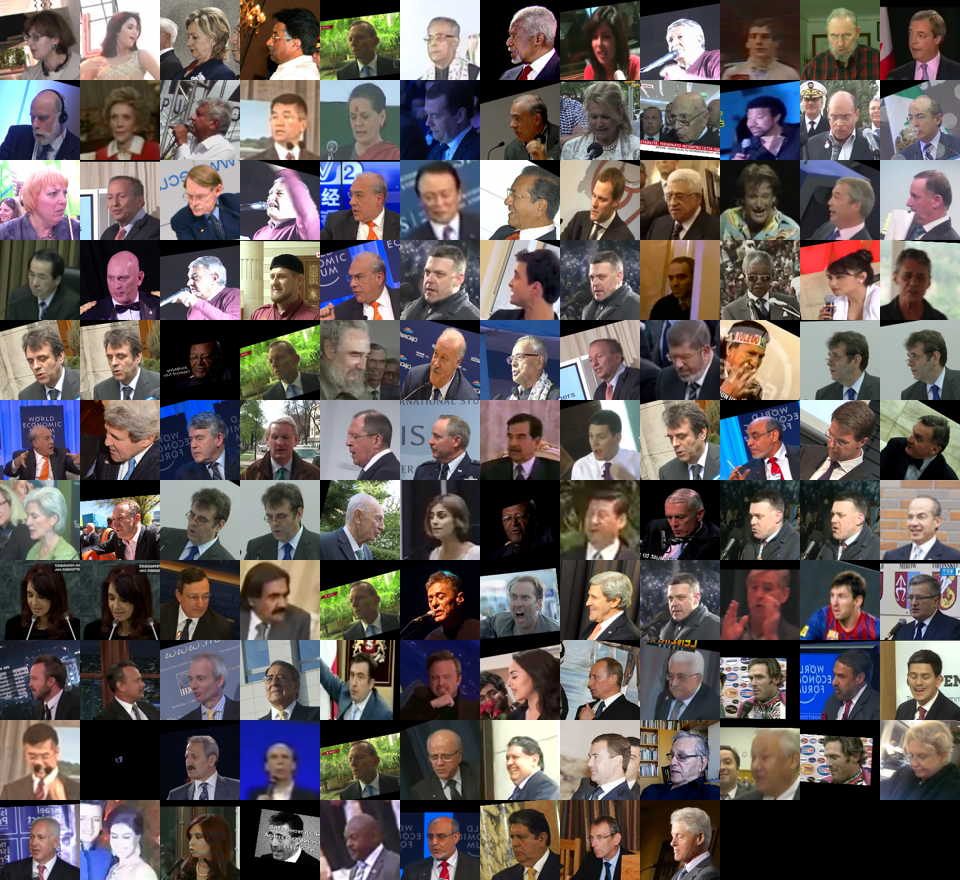}
\vskip 0.1cm
\includegraphics[width=6.0cm]{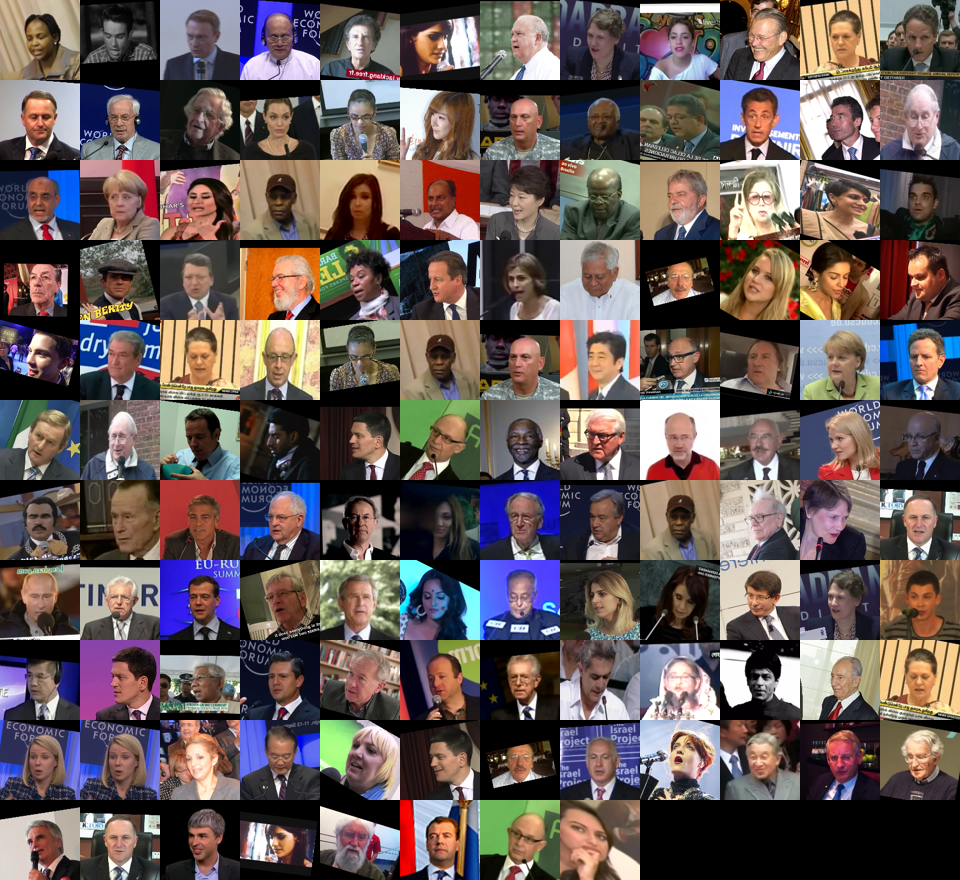}
\vskip 0.1cm
\includegraphics[width=6.0cm]{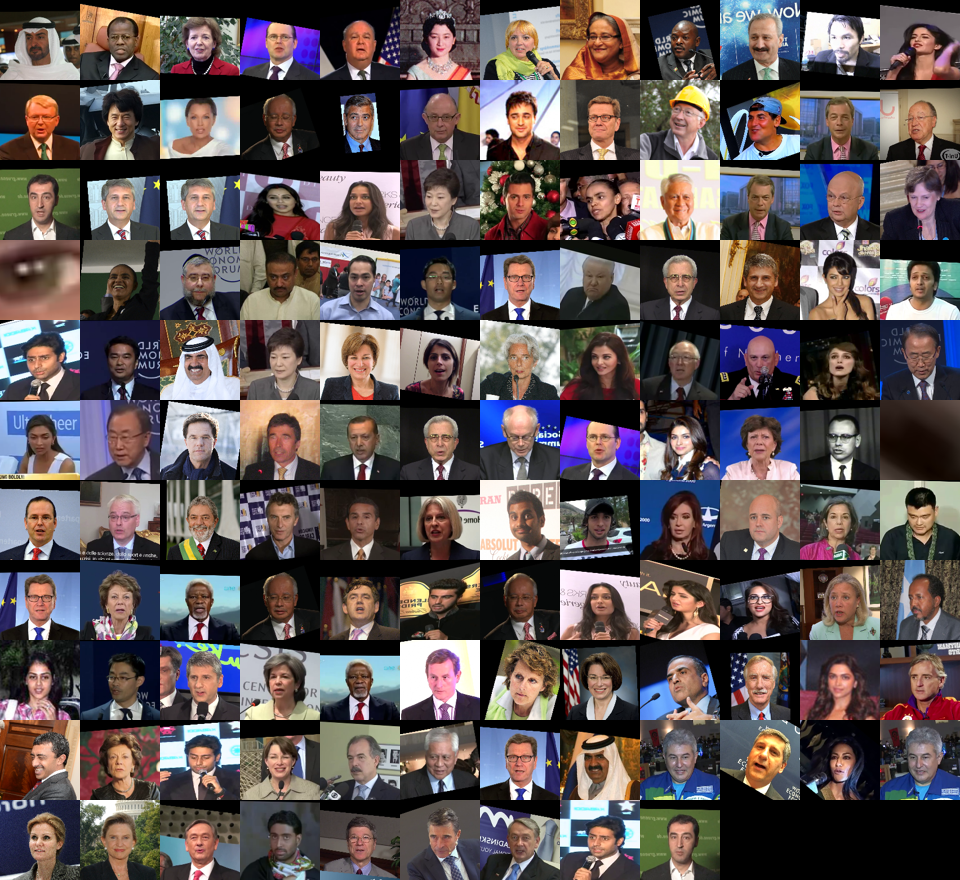}
\caption{Images (n=129) sampled at the 20\textsuperscript{th} (top), 50\textsuperscript{th} (middle),  and 90\textsuperscript{th} (bottom) percentiles of ranked distances from the origin. Face image quality seems to increase with distance from the center of the DCNN feature space.}
\label{fig:quality}
\end{figure}

Does distance from the center of the DCNN feature space index image quality? To examine this, we pulled images from different distances to the center of the space. We ranked the images according to their distance from the origin. 
Figure \ref{fig:quality} shows 258 sampled images
at the 20\textsuperscript{th}, 50\textsuperscript{th},  and 90\textsuperscript{th} percentiles of these ranked distances.  
This figure illustrates that face quality seems to increase with distance from the center of the DCNN feature space. 


\addtolength{\textheight}{-3cm}   



\section{CONCLUSIONS}

The three analyses we carried out yielded the following results. First, DCNN top-level features retain a surprising
amount of information about the original input imagery.  Yaw, pitch, and media type were readily available in the top-level DCNN codes, and could be classified with high accuracy.  

Second, in characterizing the extent to which individual features coded view-dependent or view-invariant information about faces,
we found that view-dependent coding
was a characteristic of  the \textit{identities} rather than the features.
This might imply that some identities in this dataset present with appearance characteristics  that are 
easy to detect and code across views, whereas other identities have few appearance characteristics 
that can generalize across view.  This data-dependent code is intriguing in that it suggests that DCNNs and the human visual system alike might need to exploit both types of codes to operate efficiently and accurately in unconstrained viewing conditions.
The same general finding of data-dependency held for media type as well, with some identities having consistent codes across media types
and others having disparate codes.

Finally, we found an unexpected index of image quality in the DCNN space. This took the form of distance from the origin of the space.  The clustering of poor quality images was notable in that the low quality emanated from many distinct sources. This allowed for a more generic definition of images with limited or unusable information about identity. Because low quality images cluster around the origin and quality increases with distance from the origin, we might speculate
that strong DCNN feature scores reflect robust identity information. This suggests a new method for screening out poor quality imagery in DCNNs.

In summary, a more in-depth look at DCNN features gave insight into the nature of the image information in these top-level compact feature codes.
These analyses point to data-dependent flexibility in the type of codes that emerge at the top level features, as well as the possibility of 
isolating bad data from better quality imagery.

\section{ACKNOWLEDGMENTS}

This research is based upon work supported by the Office of the Director of National Intelligence (ODNI), Intelligence Advanced Research Projects Activity (IARPA), via IARPA R\&D Contract No. 2014-14071600012. The views and conclusions contained herein are those of the authors and should not be interpreted as necessarily representing the official policies or endorsements, either expressed or implied, of the ODNI, IARPA, or the U.S. Government. The U.S. Government is authorized to reproduce and distribute reprints for Governmental purposes notwithstanding any copyright annotation thereon. 



\bibliographystyle{IEEEtran}
\bibliography{faces2}






\end{document}